\documentclass[article,12pt,onecolumn,]{IEEEtran}
\usepackage{lipsum,babel}
\IEEEoverridecommandlockouts
\usepackage{cite}
\usepackage{amsmath,amssymb,amsfonts}
\usepackage{algorithmic}
\usepackage{graphicx}
\usepackage{textcomp}
\usepackage{xcolor}
\usepackage{blindtext}
\usepackage{multicol}
\def\BibTeX{{\rm B\kern-.05em{\sc i\kern-.025em b}\kern-.08em
    T\kern-.1667em\lower.7ex\hbox{E}\kern-.125emX}}
\begin{document}

\title{Comparison of Probabilistic Deep Learning Methods for Autism Detection\\
}

\author{\IEEEauthorblockN{Godfrin Ismail}\\
\IEEEauthorblockA{\textit{Department of Computer Science} \\
\textit{Dedan Kimathi University of Technology}\\
ismail.godfrin21@students.dkut.ac.ke}\\
\and
\IEEEauthorblockN{Kenneth Chesoli}\\
\IEEEauthorblockA{\textit{Department of Computer Science} \\
\textit{Dedan Kimathi University of Technology}\\
chesoli.kenneth@dkut.ac.ke}\\
\and
\IEEEauthorblockN{Golda Moni}\\
\IEEEauthorblockA{\textit{Department of Computer Science} \\
\textit{Dedan Kimathi University of Technology}\\
moni.golda21@students.dkut.ac.ke}\\
\and
\IEEEauthorblockN{Kinyua Gikunda}\\
\IEEEauthorblockA{\textit{Department of Computer Science} \\
\textit{Dedan Kimathi University of Technology}\\
patrick.gikunda.dkut.ac.ke}
}

\maketitle

\begin{abstract}
Autism Spectrum Disorder (ASD) is one neuro developmental disorder that is now widespread in the world. ASD persists throughout the life of an individual, impacting the way they behave and communicate, resulting to notable deficits consisting of social life retardation, repeated behavioural traits and a restriction in their interests. Early detection of the disorder helps in the onset treatment and helps one to lead a normal life. There are clinical approaches used in detection of autism, relying on behavioural data and in worst cases, neuroimaging. Quantitative methods involving machine learning have been studied and developed to overcome issues with clinical approaches. These quantitative methods rely on machine learning, with some complex methods based on deep learning developed to accelerate detection and diagnosis of ASD. These literature is aimed at exploring most state-of-the-art probabilistic methods in use today, characterizing them with the type of dataset they're most applied on, their accuracy according to their novel research and how well they are suited in ASD classification. The findings will purposely serve as a benchmark in selection of the model to use when performing ASD detection.
\end{abstract}


\section{Introduction}
Autism spectrum disorder (ASD) is a term that describes a set of neuro-developmental disorders that includes autism and Asperger's syndrome (AS)\cite{b1}. Individuals with ASD tend to have a specific combination of deficits ranging from social communication impairments, recurrent behavioral traits worsened by restricted interests that begin in the early stages of their lives\cite{b2}. In a study carried out by\cite{b3} on the prevalence of ASD since 2014, the findings were that the disorder prevalence has risen at a high rate around the world, bringing about the possibility of having an autism epidemic. \cite{b3} mentions that this prevalence rate brings the need of availing ASD related services that include training professionals and having efficient methods for identification and diagnosis. The current diagnosis criteria for ASD are based on three main symptoms: impaired social communication, limited interests, and repetitive behaviors \cite{b4}. Because ASD is so complicated and varied, it can take a long time from the time behavioral indications are noticed to the time a definitive diagnosis is made \cite{b5}. This means that there is always a lag in treatment of the disorder. The prevalence rate also has brought the surge in the amount of research being carried out to aid in the detection and diagnosis of the disorder \cite{b3}.

Early diagnosis and early special education is necessary to allow a patient to learn to perform activities and lead an almost normal life \cite{b6}. \cite{b6} found that there are three types of data that can aid in diagnosis of ASD, these data can either be behavioral data, data that is based on EEG (electroencephalogram) and magnetic resonance imaging (MRI) data. A number of brain imaging investigations have revealed functional and structural abnormalities in the brains of people with ASD \cite{b7}. Magnetic resonance imaging (MRI) investigations, for example, have revealed that people with ASD exhibit an abnormal age-related brain growth trajectory in the frontal area \cite{b8} \cite{b9}, this strongly shows that functional brain imaging at a young age is critical for detecting persistent ASD problems \cite{b10}. \cite{b11} used multimodal brain imaging modalities [structural MRI, diffusion tensor imaging (DTI), and hydrogen proton magnetic resonance spectrum (1H-MRS)] to explore neural structure in the same group of people (19 adults with ASD and 18 adults with TD) and applied the fractional anisotropy decision tree (FA), radial diffusivity (RD), and cortical thickness as features to perform classification between ASD patients and typically developed patients (TD). This strategy eliminates the discrepancy that can occur when each imaging modality is used independently. \cite{b11} Atypical brain activation in response to various cognitive activities or diminished resting-state functional connectivity have been seen in some functional brain investigations (RSFC).

The majority of the time, MRI characteristic features retrieved empirically from brain scans are used to differentiate between children with ASD and those who do not \cite{b12}. However, due to the intricacy of ASD's pathogenic process and our poor understanding of it, the hidden variables associated with ASD, which can be utilized to accurately differentiate ASD from normal controls, are difficult to notice and identify simply by looking at brain pictures \cite{b10}. Because a deep-learning artificial neural network is a data-driven method, it can identify hidden properties in a large data set. And as the study by \cite{b13} observes, the models based on deep learning can automatically learn complex patterns that are hidden in a high-dimensional data and this allows the deep learning models to achieve  state-of-the-art performance in various domains, and of most relevance in this research is the object detection for image processing. This is the one aspect that is critical for Autism Detection and Diagnosis as described in this paper.

This literature provides an in-depth analysis on the available state-of-the-art probabilistic models that are used in detection and diagnosis of ASD, by a comparison of their applicability, a review on their accuracy metrics depending on how various literatures and research articles present them and finally reveal what models are efficient in what circumstances. Thus, this review is conducted to provide an answer to the following research questions:

\begin{itemize}
\item How is deep learning used in detection and diagnosis of ASD?
\item What are the deep learning models used in detection and diagnosis of ASD?
\item What is the best model that can be applied depending on the data provided?
\end{itemize}

First section of this paper introduces machine learning and its application in general disorder detection. The section tries to provide an insight on the possibility of having machine learning as a next method to be applied for early and easy disorder detection. Then we provide an introduction to deep learning and how it has been used in neuroimaging - a type of brain images that can be studied to analyze and reveal patterns of brain related disorders. Lastly, we present various models that our literature has found and a most efficient way to apply them. A conclusion of the review is presented.

\section{Methodology}

We determined research questions that would direct the review process in order to accomplish the goals in this study domain. We first collected related works done by other scholars on Deep learning methods used in ASD detection and diagnosis and then did a detailed review and analysis of the works. Collection of the papers was based on an internet search from various repositories, for researches that contained the following deep learning models; ASD-Diagnet, VGG16, deep neural network for ASD classification, Auto-ASD Network, Xception, Visual Geometry Group Network (VGG19) and NASNETMobile. The second criteria search was based on ASD detection and diagnosis. An in-depth analysis of the obtained models was done aligning to the following research questions: a) detection or diagnosis technique addressed, b) the model being used in the research, c)  the dataset used in the model implementation and d) accuracy of the model based on the metrics used by the researchers. The review then turns to a comparison of the identified models, their suitability given which kind of dataset and then finally how possibly the models could be used for better performance.

\section{Application of Machine Learning in Disorder Detection}
Machine learning is a field of research for computer scientists and Engineering, it is a branch of artificial intelligence since it allows one to extract meaningful patterns from examples, which is also a component when it comes to the intelligence of humans \cite{b14}. In the recent past, machines have demonstrated the ability to learn tasks that had previously been thought of being too complex for machines\cite{b15}. This clearly demonstrates that machine learning algorithms have great potential as components of computer-aided diagnosis as well as in the area of decision support systems.

The use of machine-learning approaches in conjunction with brain imaging data has enabled for the classification of mental states related with semantic category representation\cite{b16}. Patients with brain activation linked with schizophrenia have been observed in research of mental illness states \cite{b17} and with autism \cite{b18}. Studies using ML algorithms on ASD brain imaging data categorized patients as autistic or control based on their fMRI brain activation with up to 97 percent accuracy \cite{b19}. A pattern of brain activation linked to a psychological element was also discovered (self-representation). The pattern was almost non-existent in autistic participants, but was prevalent in typically developed (control) patients. \cite{b18}. Participants with ASD were classified in another study. \cite{b20}, In a population sample of 178 ASD and IQ-matched typically-developing participants, the authors achieved a classification accuracy of 76.67 percent.

Support vector machine (SVM) or Gaussian Naive Bayes (GNB) classifiers were used in the majority of studies that combined brain imaging and machine learning. The subjective nature of feature selection approaches for supervised machine learning systems could make cross-study comparisons difficult. A set of data used as the training dataset is assigned class labels in supervised methods; the classification of other data points (test data set) is based on the patterns observed in the training data (using the given labels). The algorithm works by classifying labels that have already been assigned to them (that is, they rely on feature selection, or feature engineering). The labels and attributes chosen are based on a priori assumptions or exploratory approaches.; As a result, they are subject to a degree of subjectivity. For example, the number of voxels utilized for brain imaging data categorization was empirically determined by examining sets of 100, 200, 400, and more voxels and determining the set size that works best for classification.

\section{Application of Deep Learning in Neuroimaging}
In \cite{b21} they used Deep Neural Network to examine brain states based on activities in the brain. They adopts task-based fMRI data from 499 patients to train an artificial neural network with two hidden layers and a softmax output layer to classify the data into seven categories linked to the tasks: Working Memory, Emotion, Gambling, Language, Motor, Relational, and Social Memory. When compared to supervised learning methods, deep models produced higher results (mean accuracy of 50.74 percent) (mean accuracy of 47.97 percent), supervised models such as Linear Regression and Support Vector Machine. In \cite{b20}, the researchers employed deep learning and structural T1-weighted images to distinguish schizophrenia patients vs healthy controls; using data from the PREDICT-HD project, the investigators also classified individuals with Huntington's disease versus healthy controls. First, they used data from four studies to classify 198 schizophrenic patients and 191 controls (WPIC). \cite{b20}. trained a three-depth Deep Belief Network (50 hidden units in the first layer, 50 in the second layer, and 100 in the top layer). They were able to reach 90\% classification accuracy utilizing features extracted from three DBMs, compared to 68 percent using raw data in a Support Vector Machine. Deep learning has a lot of potential for clinical brain imaging applications, according to the scientists.

\section{Review of the Deep Learning Models for Autism Detection}
Use of behavioral data for detection and diagnosis of ASD has proven to be difficult, requiring a move towards quantitative diagnosis driven by machine learning\cite{b22}. Magnetic Resonance (MRI) and functional Magnetic Resonance Imaging (fMRI) data is the quantitative data that has been used extensively by most researchers to build deep learning methods for ASD diagnosis and detection\cite{b22}. ASD-Diagnet is a deep learning based model used by \cite{b23} and \cite{b22}.

Most studies that have implemented ASD-Diagnet used fMRI data contained in the ABIDE dataset\cite{b23}. ASD-Diagnet is a hybrid deep learning model that is used in ASD detection based on the ABIDE dataset, containing 1035 subjects that were studied at 17 different brain imaging centers\cite{b24}. ASD-Diagnet employs an autoencoder and a single layer perceptron to classify ASD patients from those that were not, called typically developed and abbreviated as TD\cite{b22}. \cite{b22} used a cross-validation to test the accuracy of the model against other models, ASD-Diagnet obtained a maximum classification accuracy of 82\%. \cite{b22} terms this as a better performance relative to other state of the art models\cite{b24}. \cite{b24} criticizes the model as not having properly undertaken the effect of structural and phenotypic information of the subjects in creation of hybrid features. They have also noted that the model’s accuracy fluctuates when subjected to a dataset other than ABIDE.

\begin{figure}[htbp]
\centerline{\includegraphics[scale = 0.45]{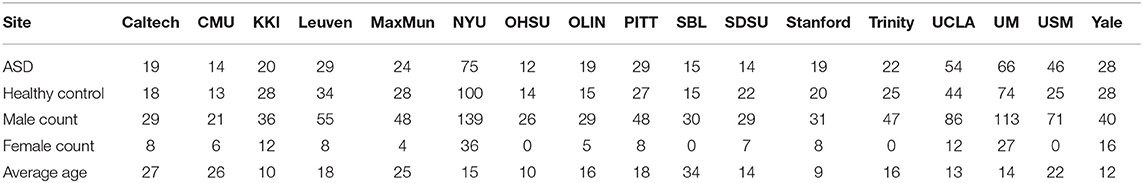}}
\caption{Class information in the ABIDE Dataset}
\label{fig}
\end{figure}

VGG16 is a state-of-the-art transfer learning based deep learning model that was pioneered by \cite{b26}. VGG16 works by applying a classification function on a dataset consisting of facial images.  \cite{b26} trains the model with two datasets; East Asia ASD children Facial Image Dataset or otherwise East Asian Dataset and Kaggle Autism Facial Dataset. The East Asian dataset contains 1122 images (with 561 for TD children) of children aged between 2 and 12 years of the same race. The Kaggle Autism Facial Dataset has 2936 images of both ASD and TD children, evenly distributed \cite{b25}.

\begin{figure}[htbp]
\centerline{\includegraphics{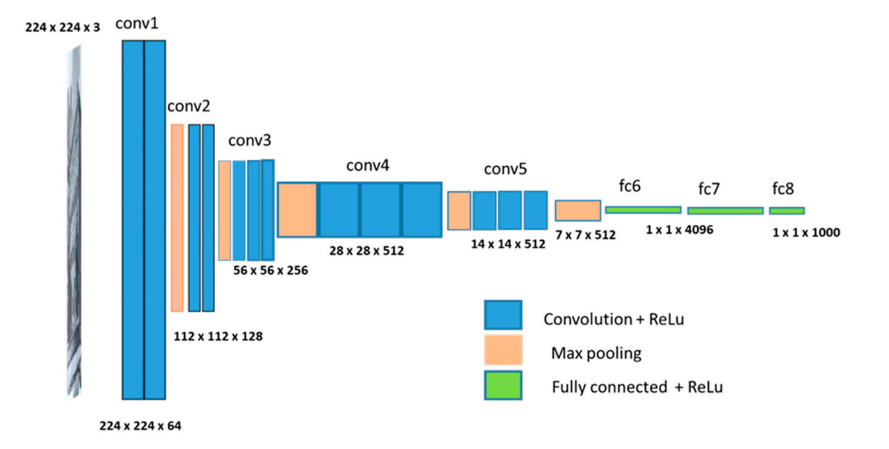}}
\caption{The Architecture of VGG16\cite{b26}}
\label{fig}
\end{figure}

In this feasibility and accuracy study, \cite{b26} used the East Asian dataset for model training and verification because the ASD facial images in this dataset are from clinically diagnosed children from a single race. The study bridges the gap of applying computer vision in ASD screening ASD in children using their facial images. The high classification accuracy of 95\% and F1-score of 0.95 obtained by the deep learning model trained with the East Asian dataset indicates that it is viable to use children’s facial images as a low-cost solution to screen for ASD to achieve early intervention objectives. Suggested that race  information be provide as a prerequisite to use the model to eliminate errors brought about from the anthropometric differences among races.

A deep neural network was used by \cite{b27} to classify ASD children from TD children. The data used by \cite{b27} is not the same dataset used by \cite{b26} in detection of ASD. \cite{b27} uses Kaggle dataset that contains descriptive features of both children with ASD and those without ASD.

The DNN developed by \cite{b27} has five layers: sequence input layer, fully connected layer, Softmax layer, Long-Short Term Memory layer and classification layer. Its architecture can is as below:

\begin{figure}[htbp]
\centerline{\includegraphics[scale = 1]{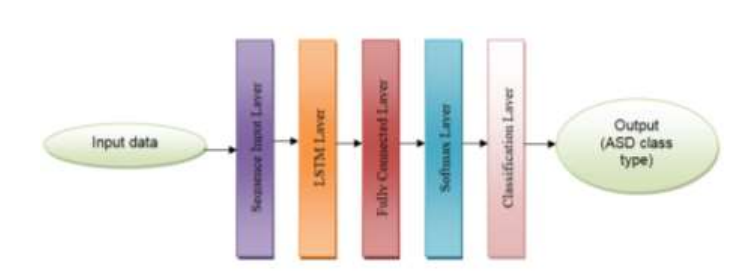}}
\caption{DNN structure created by\cite{b27}}
\label{fig}
\end{figure}

The dataset used in \cite{b27} was provided by Fadi Fayez Thabtah, who used the mobile app ASDTests to screen people of various ages for ASD, contains different category of data. The toddler data set has 18 unique items and one output class, whereas the child, adolescent, and adult data sets have 21 unique things and one output class. \cite{b27} A classifier model is created using deep learning to categorise ASD and no ASD class for the purpose of diagnosing ASD. From training data, a classifier model is created, and it is then assessed using test data. \cite{b27} It can be shown that utilising the suggested DNN model, 85.24 percent of toddler data is correctly classified, leaving 14.76 percent of the relevant data that did not accomplish their target class. Additionally, \cite{b27} discovered that the accuracy rate for adolescent data was 84.21 percent, allowing 15.79 percent of the relevant data to fall outside the target group. In the child data set, accuracy is 85.71 percent, with 14.29 percent of the data not reaching the intended class. \cite{b27} The adult data set's highest accuracy rate of 89.26\% is finally attained, with 10.74\% of the data not meeting the target class. \cite{b27}

In \cite{b22} deep learning model, utilising only functional Magnetic Resonance Imaging (fMRI) data, it proposes the Auto-ASD-Network model to distinguish between people with Autism disorder and healthy subjects. A multilayer perceptron (MLP) with two hidden layers makes up the model. \cite{b22} does data augmentation using the Synthetic Minority OverSampling (SMOTE) algorithm to create synthetic data and prevent overfitting, which improves classification accuracy. The deep learning model used the ABIDE dataset.

The \cite{b22} additionally looks into how well the deep learning model worked to extract the features. This is accomplished by using the deep learning model's hidden layer, which contains extracted features, as the input to an SVM classifier in addition to performing tests on the model as a classifier. \cite{b22} Since the best values of some SVM hyperparameters, such as the kernel function and penalty, are not known in advance, \cite{b22} employ a technique called Auto Tune Models (ATM) to automate the hyperparameter tuning procedure. The trials demonstrate a considerable increase in classification accuracy when the oversampling method and SVM are used. For four fMRI datasets, it achieves classification accuracy of more over 70\%, with the best accuracy being 80\%. \cite{b22} It enhances SVM performance by 26\%, standalone MLP performance by 16\%, and the state-of-the-art ASD classification approach by 14\%. \cite{b22}

\cite{b28} suggested the use of Xception, NASNETMobile, and VGG19, deep learning models based on transfer learning, to identify autism using face traits of autistic and typical children. Each model underwent training and testing to identify the characteristics that, based on face features, classify youngsters as autism or normal. \cite{b28} This study compared facial pictures of autistic and typical kids that were taken from the publicly available online Kaggle platform. \cite{b28} The NASNetMobile model had the lowest performance level (78\%) and the Xception model had the best testing accuracy (91\%) of the three models. \cite{b28}

The \cite{b28} employs a variety of performance evaluation metrics, including a confusion matrix, accuracy, sensitivity, and specificity, for the three pretrained models. A table containing the true and false values of the test results is represented by a confusion matrix, a sort of classification performance metric. \cite{b28} In the confusion matrix of the Xception model, the True Positives represented 132 of the 150 autistic children, the False Negatives represented 18, the True Negatives represented 141 of the 150 normally developing children, and the False Positives represented 9. \cite{b28} The VGG19 model's accuracy increased during the training phase from 56\% to 85\% after 25 epochs, then it decreased to 82\% during the validation phase. \cite{b28} The performance of NASNetMobile was subpar. 70–90\% of the time it was accurate during the training phase, and 65–78\% of the time during the validation phase. \cite{b28} The accuracy of the Xception model was 100\% on the training data and 91\% on the testing data. \cite{b28} Although the data generator obtained the dataset from online sources, which made the differences in ages and photographic quality evident, the Xception model displayed the highest level of accuracy. \cite{b28}

\section{Conclusion and Recommendation}
There is a growing interest in autism which has propagated advancement in the global-health know how and capacilties. Moreover, the number of autistic children has risen in recent years, making it imperative that attention is given to it in order to help these children get medical attention and live fulfilling lives. This is made possible through early diagnosis to help in deciding the best treatment and care to offer them. This paper evaluated the performance of various deep learning models in detecting ASD: ASD-Diagnet, VGG16, deep neural network for ASD classification, Auto-ASD Network, Xception, Visual Geometry Group Network (VGG19) and NASNETMobile. Each model was trained suing a different dataset and the classification accuracy recorded. Based on the researchers findings, VGG16 achieved the highest classification accuracy at 95\%, closely followed by Xception as 91\%. The results from the deep learning models show us the possibility of deep learning being used by specialists and families to give more accurate diagnosis of autism, especially when they are still young. More research and future work should be undertaken to design models that not only detect ASD, but can also diagnose the severity of ASD.

\vspace{12pt}

\end{document}